\documentclass[10pt,twocolumn,letterpaper]{article}

\usepackage[pagenumbers]{cvpr} % To force page numbers, e.g. for an arXiv version

\usepackage[table]{xcolor}

\usepackage{multicol}
\usepackage{multirow}
\usepackage{booktabs}
\usepackage{utfsym}
\usepackage{makecell}
\usepackage{subcaption}
\usepackage{tcolorbox}
\usepackage{amsmath}

\usepackage{tikz,pgfplots,pgfplotstable}
\usepackage{color}
\usepgfplotslibrary[groupplots]
\definecolor{mediumelectricblue}{rgb}{0.12,0.314,0.588}
\definecolor{mossgreen2}{RGB}{138,154,91}
\definecolor{internationalorange}{RGB}{100,31,0}
\definecolor{lightsalmon}{RGB}{255,160,122}

\definecolor{selfyellow}{RGB}{255,209,0}
\definecolor{selfblue}{RGB}{89,138,234}
\definecolor{selfgreen}{RGB}{133,235,133}

\usepackage{diagbox}
\usepackage{pifont}
\usepackage{appendix}

\newcommand{\cmark}{\ding{51}}  % Checkmark
\newlength{\savedwidth}
\newcommand{\whline}[1]{\noalign{\global\savedwidth\arrayrulewidth \global\arrayrulewidth #1}%
\hline \noalign{\global\arrayrulewidth\savedwidth}}

\definecolor{cvprblue}{rgb}{0.21,0.49,0.74}
\usepackage[pagebackref,breaklinks,colorlinks,citecolor=cvprblue]{hyperref}
\def\paperID{} % *** Enter the Paper ID here
\def\confName{}
\def\confYear{}
\newcommand*\samethanks[1][\value{footnote}]{\footnotemark[#1]}

\title{Revisiting MLLMs: An In-Depth Analysis of Image Classification Abilities}

\author{
Huan Liu$^{1,2}$\thanks{Equal contribution. Work done when H. Liu was an intern at Baidu.}\quad
Lingyu Xiao$^{2}$\samethanks\quad
Jiangjiang Liu$^{2}$\quad
Xiaofan Li$^{2}$\quad
Ze Feng$^{2}$\quad
Sen Yang$^{2}$\quad
Jingdong Wang$^2$\thanks{Corresponding author.}\\[1.2mm]
$^1$Beijing Jiaotong University \quad
$^2$Baidu VIS\\
{\tt\small liu.huan@bjtu.edu.cn \quad wangjingdong@baidu.com}
}

\newlength\savewidth

\begin{document}
\maketitle

\begin{abstract}
With the rapid advancement of Multimodal Large Language Models (MLLMs), a variety of benchmarks have been introduced to evaluate their capabilities. While most evaluations have focused on complex tasks such as scientific comprehension and visual reasoning, little attention has been given to assessing their fundamental image classification abilities.
In this paper, we address this gap by thoroughly revisiting the MLLMs with an in-depth analysis of image classification. Specifically, building on established datasets, we examine a broad spectrum of scenarios, from general classification tasks (e.g., ImageNet, ObjectNet) to more fine-grained categories such as bird and food classification. Our findings reveal that the most recent MLLMs can match or even outperform CLIP-style vision-language models on several datasets, challenging the previous assumption that MLLMs are bad at image classification \cite{VLMClassifier}.
To understand the factors driving this improvement, we conduct an in-depth analysis of the network architecture, data selection, and training recipe used in public MLLMs. Our results attribute this success to advancements in language models and the diversity of training data sources. Based on these observations, we further analyze and attribute the potential reasons to conceptual knowledge transfer and enhanced exposure of target concepts, respectively.
We hope our findings will offer valuable insights for future research on MLLMs and their evaluation in image classification tasks.
\end{abstract}

\section{Introduction}

\begin{figure}
    \centering
    \includegraphics[width=1\linewidth]{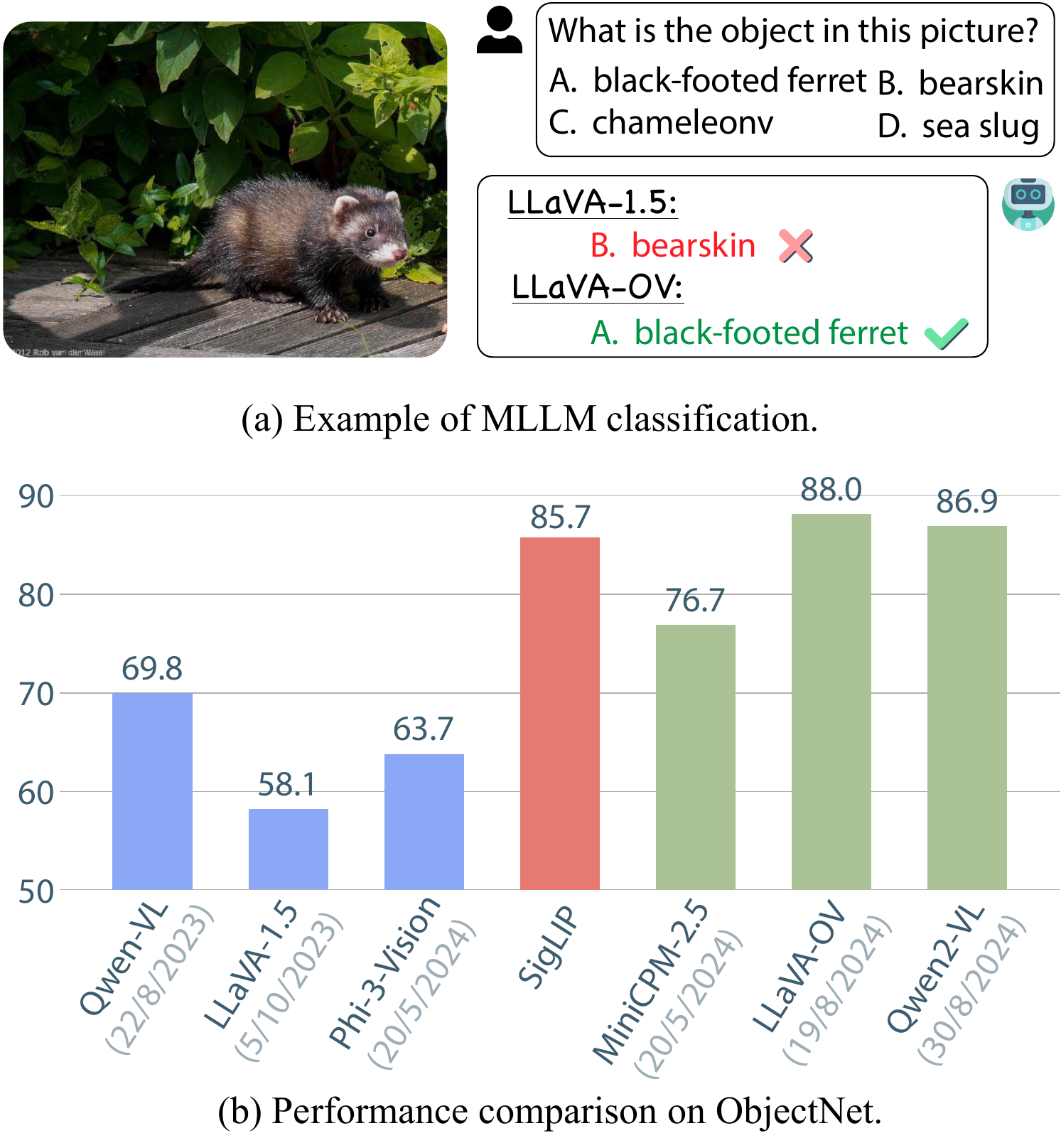}
    \vspace{-8mm}
    \caption{\textbf{Specific and overall comparisons of image classification.} (a) LLaVA-OV \cite{llava-ov} handles well on bad cases, such as ``black-footed ferret'' recognition, than previous LLaVA-1.5 \cite{llava}. (b) Recent proposed MLLMs obtain comparable or even better classification results on ObjectNet~\cite{barbu2019objectnet} dataset than SigLIP \cite{siglip}.}
    \label{fig1}
\end{figure}

With the rapid development of 
Multimodal Large Language Models (MLLMs)
in recent years, there have plenty of multimodal benchmarks \cite{hudson2019gqa,mathew2021docvqa,jin2024rwku} been proposed 
focusing on complex tasks, requiring high-level understanding and reasoning, such as 
answering engineering problems \cite{yue2023mmmu} and calculating statistical questions from charts \cite{chartqa}.
Despite increasingly comprehensive evaluation of such complex abilities in existing benchmarks, a deep investigation into MLLMs' fundamental perceptual abilities---particularly in image classification, a core task in computer vision research over the past decades---has remained conspicuously absent.

\begin{figure*}[t]
    \centering
    \includegraphics[width=1\linewidth]{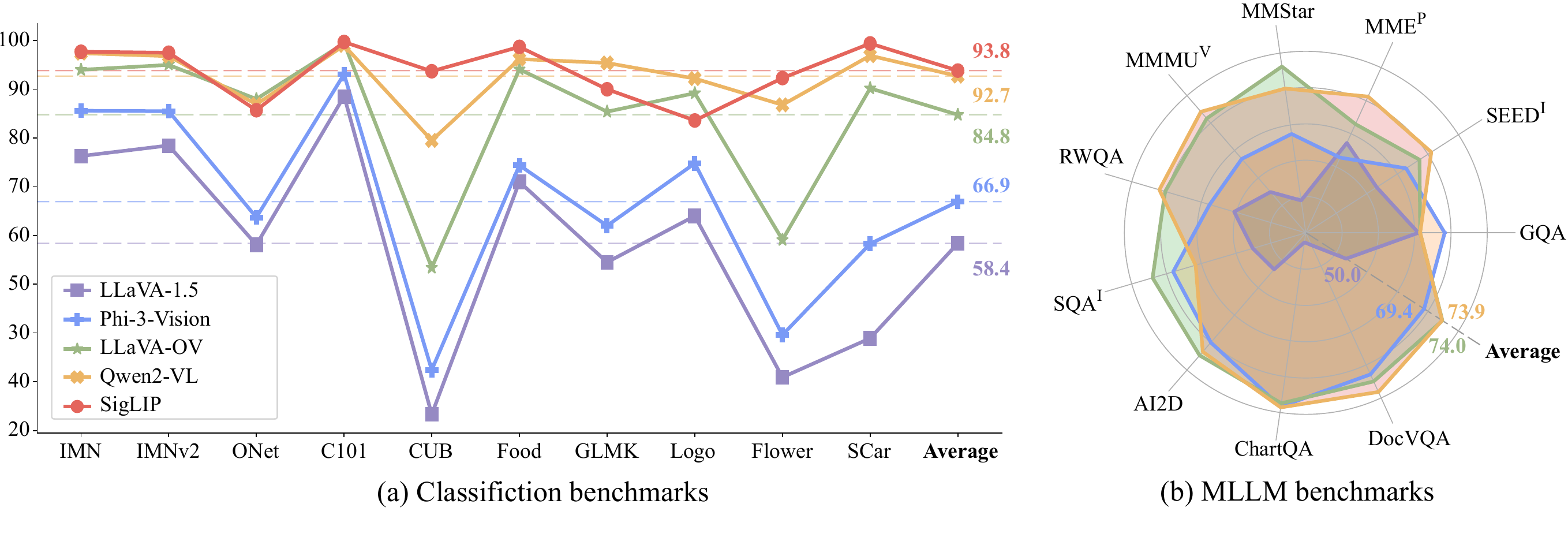}
    \vspace{-10mm}
    \caption{\textbf{Performance comparison of public MLLMs on various classification and MLLM benchmarks.} Here, we compare the established LLaVA-1.5 \cite{llava}, Phi-3-Vision \cite{phi3vision}, recent LLaVA-OV \cite{llava-ov} and Qwen2-VL \cite{qwen2vl} on a total of ten conventional image datasets (detailed in Table~\ref{tab:data_classification}), such as ImageNet~\cite{deng2009imagenet} and ObjectNet~\cite{barbu2019objectnet} for general classification evaluation, as well as CUB200~\cite{cub} and Food101~\cite{bossard2014food} for fine-grained scenarios. 
    We also report the results on ten well-established MLLM benchmarks, covering evaluation with four common categories \cite{tong2024cambrian}, including 
    general, knowledge, chart \& OCR, and vision-centric.
    Best view in color.}
    \label{fig2}
\end{figure*}

A recent study \cite{VLMClassifier} notices this problem and turns to explore the image classification ability of MLLMs with four traditional datasets, including ImageNet \cite{deng2009imagenet}, Caltech101 \cite{caltech}, Flowers-102 \cite{flower102}, and StanfordCars \cite{stanfordcars}.
They find MLLMs, such as LLaVA-1.5 \cite{llava}, significantly underperform CLIP \cite{clip,siglip} on standard image classification tasks like ImageNet and conclude that \textit{MLLMs are bad at image classification}.
Intriguingly, as shown in Figure~\ref{fig1} (a), we observe the recently proposed LLaVA-OV \cite{llava-ov} can handle previous bad cases of LLaVA-1.5 well in practice.
To further verify this phenomenon, we perform a classification evaluation based on the conventional ObjectNet~\cite{barbu2019objectnet} dataset.
The detailed structures and sizes of used MLLMs are listed in Table~\ref{tab:cls_compare_mllm_clip}, omitted for simplicity in the rest paper.
Surprisingly, as illustrated in Figure~\ref{fig1} (b), we notice that many recently-proposed public MLLMs, \emph{e.g.}, MiniCPM-V-2.5 \cite{minicpm}, LLaVA-OV \cite{llava-ov} and Qwen2-VL \cite{qwen2vl}, have caught up or even exceeded the SigLIP \cite{siglip} (an improved variant of CLIP) baseline on the ObjectNet dataset, which challenges the prior assumption.

This motivates us to perform an in-depth analysis of image classification by revisiting MLLMs, {\em i.e.}, \textit{are all MLLMs truly poor classifiers?}
Concretely, building upon ten representative datasets, we consider a wide range of aspects, covering both general and fine-grained scenarios, resulting in a robust and comprehensive classification benchmark.
Following the standard MLLM and CLIP evaluation protocols \cite{li2023seed,liu2024mmbench,clip}, we reformulate the image classification task into commonly used multiple-choice questions for MLLMs and image-text pairs for the zero-shot matching of CLIP-style models, respectively.
Our evaluation encompasses lots of publicly available MLLMs, including both established and newly released models, as well as CLIP-style models that correspond to the representative vision towers used in MLLMs.
As shown in Figure~\ref{fig2} (a), one can see that recent MLLMs, especially Qwen2-VL, can match the SigLIP baseline across various conventional image classification datasets, achieving $92.7\%$ average accuracy.
On the other hand, we notice that there are huge improvements between LLaVA-1.5 \cite{llava} and LLaVA-OV \cite{llava-ov}, which adopt a similar LLaVA framework with a projector to connect vision tower and large language model (LLM), typically over $+26.4\%$ gains in average classification.
Alongside this, we also observe consistent improvement ($50.0$ of LLaVA-1.5 {\em v.s.} $74.0$ of LLaVA-OV) within previous well-established MLLM benchmarks in Figure~\ref{fig2} (b).
Driven by these observations, we aim to investigate the key factors underlying this improvement, in the advancement of MLLMs, \textit{what makes an MLLM a good image classifier?}

To answer this question, we investigate the core design choices in terms of MLLM's network architecture ({\em e.g.}, different combinations of vision towers and LLMs), data selection ({\em e.g.}, training data in different MLLM training stages), and training recipe ({\em e.g.}, parameters' tunability of MLLM and image high-resolution strategy).
Building upon extensive analyses, we conclude both better LLMs and diverse training data are able to enhance the image classification ability of MLLM, particularly the former, while the training recipe only has a slight impact.

To understand the rationale, we further analyze the central role that LLMs and diverse training data play in enhancing image classification abilities, that is \textit{why do the LLM and data matter?}
For the LLMs part, we explore how the conceptual knowledge embedded in different LLMs, drawn from their respective pre-training corpora, influences image classification.
Specifically, given an image, we leverage a privileged MLLM to identify the object class and generate supporting justifications without explicitly naming the class.
These justifications are then provided to LLMs, prompting them to choose a specific class from other confusing categories.
We find that LLMs in MLLMs with better classification performance also achieve better results in this text-only LLM testing.
For the training data part, we observe that under the same network architecture, training with larger-scale data, such as from LLaVA-$665$K \cite{llava} to SI-$3.2$M \cite{llava-ov} in supervised fine-tuning, results in consist classification improvements.
Thus, taking food as a specific domain, we gather relevant data within SI-$3.2$M and incorporate it with LLaVA-$665$K to form an enhanced domain-specific dataset to re-perform supervised fine-tuning of LLaVA-1.5.
An obvious increase in food-related classes can be observed during evaluation.
Finally, we attribute the potential reasons to better conceptual knowledge transfer for LLMs and enhanced exposure of target concepts for diverse training data, respectively.

Data and code will be available and we hope our findings can provide insights for further research on both MLLM and its image classification evaluations.

\section{Related Works}
\subsection{Image Classification}
Image classification is a foundational task within the domain of computer vision. Early approaches to this problem employed machine learning algorithms such as SVMs and MLPs, focusing on widely recognized benchmarks like MNIST~\cite{lecun1998mnist} and CIFAR~\cite{krizhevsky2009learning}. The landscape of image classification underwent a significant transformation with the introduction of the large-scale dataset ImageNet~\cite{deng2009imagenet}. The advent of deep learning, particularly through AlexNet~\cite{alexnet}, marked a pivotal breakthrough by achieving substantial improvements over traditional machine learning methods on ImageNet. This success spurred the development of increasingly sophisticated CNN architectures, such as VGGNet~\cite{vgg}, InceptionNet~\cite{InceptionNet}, and ResNet~\cite{resnet}, which enhanced model capacity for handling expansive datasets and deeper network structures without succumbing to overfitting. The field has recently seen a paradigm shift with the emergence of vision transformers~\cite{alexey2020image}, leading to the dominance of transformer-based methods~\cite{liu2021swin,liu2022swinv2,zhai2022scaling,dai2021coatnet} characterized by their scalability in terms of parameters and data. Inspired by these developments, recent works~\cite{clip,eva02,dfn} have integrated language description into self-supervised pre-training frameworks, achieving state-of-the-art performance on ImageNet~\cite{deng2009imagenet}. 

Nevertheless, within the context of MLLMs, the specific performance of image classification tasks remains an open question that has yet to be thoroughly explored.

\subsection{MLLM Evaluation}
The capabilities of MLLMs can be categorized into three primary dimensions: perception, comprehension, and reasoning \cite{gaur2024detect}. To evaluate these capabilities, a diverse array of benchmarks and methodologies have been developed~\cite{liu2024mmbench,fu2023mme,yue2023mmmu,llava}. Notably, benchmarks such as those in~\cite{li2023seed,liu2024mmbench} encompass over 12 evaluation dimensions, while others like~\cite{yue2023mmmu} offer comparisons across more than 30 subjects. Despite these efforts, recent studies~\cite{mmstar,tong2024cambrian} suggest that performance on certain benchmarks~\cite{lu2022sqa,yue2023mmmu,kembhavi2016ai2d} is influenced more by comprehension and reasoning than by visual perception, thereby complicating the assessment of MLLM perception capabilities. Although some benchmarks~\cite{mmstar,fu2023mme} aim to highlight the visual grounding abilities of MLLMs, fundamental visual tasks such as image classification remain conspicuously underexplored. 

The work most closely related to our study is~\cite{VLMClassifier}, which notices the challenges that MLLMs face in image classification. However, their findings tend to be somewhat self-evident and are limited by the scope of benchmarks and the diversity of MLLMs examined. 
This paper addresses these issues by comprehensively performing an in-depth analysis of image classification for MLLMs.

\begin{table}[tp]
\centering
\small
\renewcommand{\arraystretch}{1.1}
\setlength{\tabcolsep}{1.2pt}
\caption{\textbf{The datasets used for image classification benchmark.} We provide the abbreviated name (`Abbr.') for each dataset used in the rest tables. `*': we filter out the test samples that lack grounding truth labels. Here, metrics are reported on \texttt{val} set for IMN, matched-frequency \texttt{test} set for IMNv2 and \texttt{test} set for others.}
\label{tab:data_classification}
\begin{tabular}{l|l|l|r|r}
Type & Dataset & Abbr. & \#Samples  & \#Classes \\ \whline{1pt}
\multirow{4}{*}{General}
& ImageNet~\cite{deng2009imagenet} & IMN & 50K & 1,000 \\ 
& ImageNetv2~\cite{imagenetv2} & IMNv2 & 10K & 1,000 \\
& ObjectNet~\cite{barbu2019objectnet} & ONet & 50K & 313 \\
& Caltech101~\cite{caltech} & C101 & 4,331 & 101 \\ \hline
\multirow{6}{*}{\shortstack{Fine-\\grained}} 
& CUB200~\cite{cub} & CUB & 5,794 & 200 \\
& Food101~\cite{bossard2014food} & Food & 25K & 101 \\
& Google Landmarks v2*~\cite{weyand2020google} & GLMK & 1,655 & 709 \\
& FlickrSportLogos-10~\cite{logo} & Logo & 500 & 10 \\
& Flowers102~\cite{flower102} & Flower & 6,149 & 102 \\
& StanfordCars~\cite{stanfordcars} & SCar & 8,041 & 196 \\
\end{tabular}
\end{table}

\section{Are All MLLMs Truly Poor Classifiers?}
\label{sec:mllms are bad}
In this section, we address the critical question: are all MLLMs truly poor classifiers?
We commence by outlining the benchmarks and classification evaluation protocols employed, followed by presenting the experimental results across various categories of MLLMs, as well as CLIP-style vision-language models. 
Our findings indicate that while some MLLMs exhibit lower performance compared to traditional CLIP-style models, the latest MLLMs have reached performance levels that are comparable to these models.

\subsection{Benchmarks}
Unlike the limited evaluation scenarios in \cite{VLMClassifier}, we conduct a comprehensive evaluation from conventional image classification benchmarks alongside widely recognized MLLM benchmarks to assess the MLLMs' capabilities and relations in both low-level perception and high-level reasoning.

\begin{table*}[tp]
\centering
\footnotesize
\setlength{\tabcolsep}{2.3pt}
\renewcommand{\arraystretch}{1.2}
\caption{\textbf{Detailed accuracy comparisons between public MLLMs and CLIP-style models on the image classification benchmarks.} We provide the average (\textbf{Avg.}) accuracy in \textbf{bold}. `L' and `W' in the `text encoder' column represent the layer number and width of the corresponding text transformer encoder. 
The best MLLMs with available or unavailable code are marked as \colorbox{lightgray!20}{gray}.
\emph{We only consider MLLMs between 4B-10B to balance cost and performance.}}
\label{tab:cls_compare_mllm_clip}
\begin{tabular}{l|l|l|l|l|ccccr|ccccccr}
\multicolumn{2}{l|}{\multirow{2}{*}{~~~~~Model}} & \multirow{2}{*}{Size} & \multirow{2}{*}{Vision Tower} & Text Encoder / & \multicolumn{5}{c|}{General Classification} & \multicolumn{7}{c}{Fine-grained Classification} \\ \cline{6-17}
\multicolumn{2}{c|}{} & & & LLM & IMN & IMNv2 & ONet & C101 & \textbf{Avg.} & CUB & Food & GLMK & Logo & Flower & SCar & \textbf{Avg.} \\
\whline{1pt}
\multirow{6}{*}{\rotatebox{90}{CLIP-style}} 
& OpenAI CLIP \cite{clip}   & 0.4B & ViT-L        & L-12, W-768  & 96.3 & 95.5 & 81.8 & 98.4 & \textbf{93.0} & 88.5 & 97.9 & 90.9 & 63.6 & 78.6 & 94.3 & \textbf{85.6} \\
& EVA-CLIP-02 \cite{eva02}  & 0.4B & ViT-L        & L-12, W-168  & 96.8 & 96.9 & 86.9 & 99.3 & \textbf{95.0} & 90.3 & 98.0 & 90.2 & 65.0 & 87.1 & 98.8 & \textbf{88.2} \\ 
& DFN-CLIP \cite{dfn}      & 1.0B & ViT-H        & L-24, W-1024 & 97.6 & 97.7 & 83.3 & 99.5 & \textbf{94.5} & 96.1 & 98.7 & 95.6 & 66.4 & 93.7 & 99.6 & \textbf{91.7} \\ 
& OpenCLIP \cite{openclip}  & 1.2B & ConvNeXT-XXL & L-24, W-1024 & 96.3 & 96.4 & 83.3 & 99.4 & \textbf{93.9} & 94.6 & 97.4 & 91.2 & 77.6 & 87.1 & 99.3 & \textbf{91.2} \\ 
& OpenCLIP \cite{openclip}  & 2.5B & ViT-bigG     & L-32, W-1280 & 96.6 & 96.6 & 83.2 & 99.4 & \textbf{93.9} & 94.7 & 97.7 & 92.0 & 72.8 & 89.0 & 99.3 & \textbf{90.9} \\ 
& SigLIP \cite{siglip}     & 0.9B & ViT-SO400M   & L-27, W-1151 & 97.7 & 97.5 & 85.7 & 99.7 & \textbf{95.1} & 93.7 & 98.7 & 90.0 & 83.6 & 92.3 & 99.4 & \textbf{93.0} \\ 
\hline 
\multirow{11}{*}{\rotatebox{90}{MLLM}} 
& Qwen-VL~\cite{qwenvl} & 9.6B & ViT-bigG  & Qwen-7B & 87.5 & 88.2 & 69.8 & 95.1 & \textbf{85.2} & 43.7 & 85.1 & 74.4 & 82.8 & 63.1 & 61.6 & \textbf{68.5} \\ 
& LLaVA-1.5~\cite{llava} & 7.2B & ViT-L  & Vicuna-7B-v1.5 & 76.3 & 78.4 & 58.1 & 88.5 & \textbf{75.3} &23.3 &71.0  &54.5  &64.0  &30.9  &38.9 & \textbf{47.1} \\ 
& LLaVA-1.6~\cite{llavanext} & 7.1B & ViT-L  & Vicuna-7B-v1.5 & 81.7 & 84.5 & 66.2 &92.1 & \textbf{81.1} & 18.6 & 68.2 & 52.1 & 70.2 & 24.5 & 43.7 & \textbf{46.2} \\ 
& DeepSeek-VL~\cite{deepseekvl} & 7.3B & SigLIP-L\&SAM-B  & DeepSeek-7B & 91.6 &91.3 &80.2 &97.2 & \textbf{90.1} & 37.2 & 89.4 & 71.2 & 92.4 &56.9 &73.6 & \textbf{70.1} \\ 
& Phi-3-Vision~\cite{phi3vision} & 4.2B & ViT-L & Phi-3-Mini & 85.6      &85.5 &63.7 &93.1 & \textbf{82.0} &32.4 &74.4 &62.0 &74.8 &39.6 &58.3 & \textbf{56.9} \\
& MiniCPM-2.5~\cite{minicpm} & 8.5B & SigLIP-SO400M & Llama3-8B & 94.5      &94.0 &76.7 &98.3 & \textbf{90.9} &51.5 &88.4 &87.1 &95.0 &70.6 &82.1  & \textbf{79.1} \\ 
& InternVL-2~\cite{internvl, internvl1.5} & 8.1B & InternViT-300M & InternLM2.5-7B & 91.9 &91.2 &69.5 &97.0 & \textbf{87.4} & 51.1 & 84.5 & 79.3 & 78.6 &48.3 &62.2 & \textbf{67.3} \\ 
& MiniCPM-2.6~\cite{minicpm} & 8.1B & SigLIP-SO400M & Qwen2-7B & 95.5 & 94.3 &81.0 & 98.3 & \textbf{92.3} & 51.4 & 89.8 & 85.6 & 89.6 & 66.3 & 85.8 & \textbf{78.1} \\ 
\rowcolor{lightgray!20}
\cellcolor{white} & LLaVA-OV~\cite{llava-ov} & 8.0B & SigLIP-SO400M & Qwen2-7B & 94.0 & 95.1 &88.0 &99.2  & \textbf{94.1} &53.9  &94.1  &85.4  &89.2  &59.1  &90.2 & \textbf{78.7} \\ 
\rowcolor{lightgray!20}
\cellcolor{white} & Qwen2-VL~\cite{qwen2vl} & 8.3B & DFN-ViT-675M & Qwen2-7B & 97.4 & 96.8 &86.9 & 99.0 & \textbf{95.0} & 79.5 & 96.2 & 95.4 & 92.2 & 86.8 & 96.9 & \textbf{91.2} \\ 
\end{tabular}
\end{table*}

\noindent\textbf{Image Classification Benchmarks.}
Building upon existing representative datasets, this paper covers a wide range of classification aspects.
Specifically, a total of $10$ conventional image datasets are considered, including ImageNet~\cite{deng2009imagenet}, ImageNetv2~\cite{imagenetv2}, ObjectNet~\cite{barbu2019objectnet} and Caltech101~\cite{caltech} for the general image classification, as well as CUB200~\cite{cub}, Food101~\cite{bossard2014food}, Google Landmarks v2~\cite{weyand2020google}, FlickrSportLogos-10~\cite{logo}, Flowers102~\cite{flower102} and StanfordCars~\cite{stanfordcars} for various fine-grained scenarios.
The dataset details and abbreviated names are summarized in Table~\ref{tab:data_classification}.

\noindent\textbf{MLLM Benchmarks.}
Following \cite{tong2024cambrian}, we select $10$ widely recognized MLLM benchmarks to  comprehensively reflect the MLLMs' capabilities across four important aspects: 
\textit{General}: GQA~\cite{hudson2019gqa}, $\text{SEED}^\text{I}$~\cite{li2023seed}, $\text{MME}^\text{P}$~\cite{fu2023mme}, 
\textit{Knowledge}: $\text{MMMU}^\text{V}$~\cite{yue2023mmmu}, $\text{SQA}^\text{I}$~\cite{lu2022sqa}, AI2D~\cite{kembhavi2016ai2d}, 
\textit{Chart \& OCR}: ChartQA~\cite{chartqa}, DocVQA~\cite{mathew2021docvqa} 
and \textit{Vision-Centric}: MMStar~\cite{mmstar}, RealWorldQA (RWQA)~\cite{jin2024rwku}.
The calculation of evaluation metrics and design of prompts are aligned with previous works~\cite{tong2024cambrian,llava}.

\subsection{Classification Evaluation Protocols}
\label{sec:cls_eval_prot}

\noindent\textbf{For MLLMs.}
Following the well-established MLLM benchmarks \cite{li2023seed,liu2024mmbench}, we create a new image classification benchmark by reformulating the conventional image classification datasets into multiple-choice questions (MCQ).
To increase the difficulty of evaluation, unlike typically used MCQ with $4$ choices in \cite{li2023seed}, we extend each question to $26$ choices, corresponding from `A' to `Z' in the alphabet, with only one correct answer.
Specifically, we associate the correct label with one randomly chosen option letter, while randomly selecting the incorrect answers for the remaining $25$ options.
For each dataset, we generate $10$ semantically identical templates using GPT-4o~\cite{achiam2023gpt}, while randomly selecting one as the actual question template for each image sample.
For instance, in reformulating the CUB200~\cite{cub} benchmark, a sample question with candidate options can be as follows: 
\begin{quotation}
\noindent \textit{\textbf{\textcolor{red}{\textless image\textgreater}}
What species is the bird in this photo? 
Answer with the option’s letter from the given choices directly.\textbackslash n
A. \textbf{\textcolor{blue}{\textless class name A\textgreater}}\textbackslash n
B. \textbf{\textcolor{blue}{\textless class name B\textgreater}}\textbackslash n
$\cdots$
Z. \textbf{\textcolor{blue}{\textless class name Z\textgreater}} 
}
\end{quotation}
where \textit{\textcolor{red}{image}} represents the image for classifying and \textit{\textcolor{blue}{class name A ... Z}} are the candidate classes with one correct answer.
Comparisons and benefits with other evaluation formulations can be found in the supplementary materials.

\begin{table}[tp]
\centering
\scriptsize
\setlength{\tabcolsep}{.8pt}
\renewcommand{\arraystretch}{1.4}
\caption{\textbf{Detailed comparisons between public MLLMs on the MLLM benchmarks.}
The metric calculation is consistent with previous works \cite{tong2024cambrian,llava}.
Notations follow Table~\ref{tab:cls_compare_mllm_clip}.}
\label{tab:mllm_compare}
\begin{tabular}{l|ccccccccccr}
Model & \rotatebox{65}{GQA} & \rotatebox{65}{$\text{SEED}^\text{I}$} & \rotatebox{65}{$\text{MME}^\text{P}$} & \rotatebox{65}{MMStar} & \rotatebox{65}{$\text{MMMU}^\text{V}$} & \rotatebox{65}{RWQA} & \rotatebox{65}{$\text{SQA}^\text{I}$} & \rotatebox{65}{AI2D} & \rotatebox{65}{ChartQA} & \rotatebox{65}{DocVQA} & \textbf{Avg.} \\ 
\whline{1.pt}
Qwen-VL \cite{qwenvl} &  57.5 & 64.8 & 1549.0 & 34.5 & 37.0 & 49.3 & 68.8 & 63.0 & 65.7 & 66.3 & \textbf{58.4} \\ 
LLaVA-1.5 \cite{llava} & 62.0 & 66.2 & 1507.2 & 33.8 & 36.3 & 55.6 & 69.6 & 55.2 & 18.2 & 28.1 & \textbf{50.0} \\ 
LLaVA-1.6 \cite{llavanext} & 64.2 & 70.2 & 1509.7 & 37.7 & 35.8 & 57.9 & 70.3 & 65.3 & 54.6 & 74.4 & \textbf{60.6} \\ 
DeepSeek-VL\cite{deepseekvl} & 61.2 & 70.1 & 1482.8 & 40.5 & 38.3 & 54.2 & 80.9 & 65.3 & 59.1 & 47.0 & \textbf{59.1} \\ 
Phi-3-Vision \cite{phi3vision} & 64.9 & 71.7 & 1450.5 & 47.7 & 42.2 & 59.4 & 90.5 & 77.2 & 81.3 & 87.0 & \textbf{69.4} \\ 
MiniCPM-2.5\cite{minicpm} & 56.2 & 72.3 & 1637.2 & 51.8 & 45.8 & 63.5 & 89.2 & 78.4 & 73.0 & 86.0 & \textbf{69.8} \\ 
InternVL-2 \cite{internvl, internvl1.5} & 62.7 & 76.0 & 1624.2 & 59.4 & 48.0 & 64.4 & 97.0 & 82.3 & 82.5 & 91.9 & \textbf{74.5} \\
MiniCPM-2.6\cite{minicpm} & 56.4 & 74.0 & 1655.6 & 54.6 & 49.8 & 64.4 & 93.9 & 77.5 & 78.6 & 91.4 & \textbf{72.3} \\ 
\rowcolor{lightgray!20}
LLaVA-OV\cite{llava-ov} & 62.2 & 74.2 & 1579.7 & 61.8 & 49.4 & 66.3 & 95.9 & 81.1 & 80.0 & 90.2 & \textbf{74.0} \\ 
\rowcolor{lightgray!20}
Qwen2-VL \cite{qwen2vl} & 62.2 & 76.4 & 1686.1 & 57.2 & 50.6 & 67.2 & 84.6 & 80.0 & 81.6 & 95.0 & \textbf{73.9} \\
\end{tabular}
\end{table}

\noindent\textbf{For CLIP-style Models.}
Following the standard CLIP evaluation protocol \cite{clip}, we utilize zero-shot vision-language matching by selecting the class with the highest cosine similarity between the encoded image feature and text template features. 
The text template features are derived by 
encoding and averaging a series of shared text prompts, such as:
\begin{quotation}
\noindent\textit{A pixelated photo of the \textbf{\textcolor{blue}{\textless class name\textgreater}}. \\
A sculpture of the \textbf{\textcolor{blue}{\textless class name\textgreater}}. \\
A bright photo of the \textbf{\textcolor{blue}{\textless class name\textgreater}}. \\
$\cdots$
}
\end{quotation}
where \textit{\textcolor{blue}{class name}} will be replaced by the current candidate class.
Note that the candidate classes are the same as the ones for MLLMs. More details of text templates can be found in \cite{clip}.

\subsection{MLLMs and CLIP-style Models}
We select $6$ series of public, widely-used, state-of-the-art multimodal large language models, including Qwen-VL series \cite{qwenvl,qwen2vl}, LLaVA series \cite{llava,llavanext,llava-ov}, MiniCPM-V series \cite{minicpm}, InternVL series \cite{internvl,internvl1.5}, Phi-Vision series \cite{phi3vision}, as well as DeepSeek-VL \cite{deepseekvl}.
Due to limited computing resources, we focus on collecting MLLMs within the size range of $4$B to $10$B parameters to balance cost and performance, resulting in a total of $10$ established and newly released models, consisting of Qwen-VL, LLaVA-1.5, LLaVA-1.6, DeepSeek-VL, Phi-3-Vision, MiniCPM-V-2.5, InternVL-2, MiniCPM-V-2.6, LLaVA-OV and Qwen2-VL.

For the selection of CLIP-style vision-language models, we review the representative vision towers utilized by state-of-the-art MLLMs, obtaining a total $6$ types of vision-language models, {\em e.g.}, OpenAI CLIP \cite{clip}, EVA-CLIP-02 \cite{eva02}, DFN-CLIP \cite{dfn}, OpenCLIP ConvNeXT \cite{openclip}, OpenCLIP BigG \cite{openclip} and SigLIP \cite{siglip}.

\subsection{Results}
As illustrated in Table~\ref{tab:cls_compare_mllm_clip}, we conduct a comparative evaluation between various publicly available MLLMs and corresponding CLIP-style vision-language models on the proposed image classification benchmarks.
We report the specific accuracy on each of $10$ datasets, as well as the average performance on general and fine-grained aspects.
It is evident that \textbf{most MLLMs perform less effectively than those CLIP-style models in classification, particularly the fine-grained scenario}.
For example, LLaVA-1.5 \cite{llava} achieves an average accuracy of only $75.3\%$ and $47.1\%$ on general and fine-grained classification, whereas DFN-CLIP \cite{dfn} attains $94.5\%$ and $91.7\%$.
Thus, previous research~\cite{VLMClassifier} claims that MLLMs are bad at image classification.
However, our experiments reveal that \textbf{some recently-developed MLLMs significantly achieve progress on this task and even perform comparably to CLIP-style models.}
For instance, Qwen2-VL obtains $95.0\%$ general classification accuracy and $91.2\%$ for the fine-grained one, exceeding OpenAI CLIP by $+2.0\%$ and $+5.6\%$, respectively. 
LLaVA-OV easily beats LLaVA-1.5, which has a shared LLaVA framework, by $+18.8\%$ and $+31.6\%$ average improvements in these two classification aspects.
Besides, we also observe consistent advancements, such as $+24.0pts$ from LLaVA-1.5 to LLaVA-OV, on the well-established MLLM benchmarks, as shown in Table~\ref{tab:mllm_compare}.
Especially considering that these MLLMs are not targeted at image classification~\cite{internvl,internvl1.5}, what factors contribute to the effectiveness of MLLMs as classifiers?

\section{What Makes MLLMs Good Classifiers?}
Given the huge advancement between previously and recently proposed MLLMs on the image classification benchmarks, this section further aims to identify the underlying key factors.
We first present well-founded reasons for selecting LLaVA-1.5 \cite{llava} and LLaVA-OV \cite{llava-ov} as the models to be investigated in Section~\ref{sec:what makes good clser-motivate}.
By carefully comparing the differentiators between LLaVA-1.5 and LLaVA-OV, we summarize and explore three core design choices regarding the architecture design (in Section~\ref{sec:what makes good clser-net}), data selection (in Section~\ref{sec:what makes good clser-data}), and training recipe (in Section~\ref{sec:what makes good clser-recipe}).

\begin{table}[tp]
\centering
\small
\setlength{\tabcolsep}{1pt}
\renewcommand{\arraystretch}{1.3}
\caption{\textbf{Main differences between LLaVA-1.5 \cite{llava} \emph{vs.} LLaVA-OV \cite{llava-ov}} on LLM, vision tower and training stages with corresponding data and tunable parameters.}
\label{tab:model_diffs}
\begin{tabular}{l|c|c|c|c|c}
Model & \multicolumn{2}{c|}{LLaVA-1.5 \cite{llava}} & \multicolumn{3}{c}{LLaVA-OV \cite{llava-ov}} \\ \whline{1pt}
LLM & \multicolumn{2}{c|}{Vicuna-7B-v1.5} & \multicolumn{3}{c}{Qwen2-7B-Instruct} \\
Vision & \multicolumn{2}{c|}{CLIP-ViT-Large} & \multicolumn{3}{c}{SigLIP-SO400M} \\ \whline{1pt}
Stages & 1 & 2 & 1 & 1.5 & 2 \\ \hline
Data & LCS-558K & LLaVA-665K & LCS-558K & Mid-4M & SI-3.2M \\
Tunable & MLP & MLP+LLM & MLP & Full & Full \\
\end{tabular}
\end{table}

\subsection{Motivation}
\label{sec:what makes good clser-motivate}
In this section, the majority of our experiments are conducted using the LLaVA-1.5 \cite{llava} and LLaVA-OV \cite{llava-ov} models. 
This is mainly because of two reasons: i) Both of them belong to the LLaVA series with a similar LLaVA framework, while a significant performance disparity (up to $18.8\%$) is observed in Table~\ref{tab:cls_compare_mllm_clip}.
ii) The availability of training code and data ensures the accuracy of reproduction and facilitates further studies, especially considering that LLaVA-OV is the best-performing model among other code-available ones in our classification benchmarks. 

We list the primary differences between these two MLLMs in Table~\ref{tab:model_diffs}, involving LLM, vision tower, training data, as well as tunable settings of model parameters.
For example, LLaVA-OV incorporates Qwen2-7B-Instruct \cite{yang2024qwen2}, a more advanced LLM, than Vicuna-7B-v1.5 \cite{zheng2023judging} adopted by LLaVA-1.5.
Similarly, the vision tower in LLaVA-OV, SigLIP-SO400M \cite{siglip}, also surpasses the corresponding OpenAI CLIP ViT-Large \cite{clip} in the loss function, optimized architecture, input resolution and pre-training data.
Details can be found in \cite{siglip}.
Besides, LLaVA-OV introduces an additional high-quality knowledge-learning phase, denoted as Stage-1.5, by utilizing a $4$ million data mixture, {\em i.e.}, Mid-4M\footnote{\mbox{https://huggingface.co/datasets/lmms-lab/LLaVA-OneVision-Mid-Data}}, comprising re-captioned detailed description data, document/OCR data and Chinese language data.
Moreover, during different stages of supervised fine-tuning, LLaVA-OV has a larger volume of high-quality data from diverse sources ({\em e.g.}, SI-3.2M and OV-1.6M\footnote{https://huggingface.co/datasets/lmms-lab/LLaVA-OneVision-Data}), more tunable model parameters and higher image input resolution, compared to LLaVA-1.5.
To sum up, we categorize these differences into three aspects: network architecture ({\em e.g.}, different combinations of vision towers and LLMs), data selection ({\em e.g.}, training data in different MLLM training stages), and training recipe ({\em e.g.}, parameters' tunability of MLLM and image high-resolution strategy), investigated in subsequent sections.

\begin{table}[tp]
\centering
\small
\caption{\textbf{Ablation on the choice of network architecture.}
In the `Model' column, `L-' represents LLaVA-1.5 training protocol, and `C', `S', `V' and `Q' denote CLIP, SigLIP, Vicuna-1.5 and Qwen2, respectively.
\textit{Due to the test samples in MLLM benchmarks (i.e., $\text{SQA}^\text{I}$, AI2D, ChartQA, DocVQA) being leaked to LLaVA-OV's training data \cite{llava-ov}, we calculate the average MLLM scores w.r.t. two factors: 
Avg.$^\dag$ for the 4 leaked ones and 
Avg.$^*$ for the rest.}}
\label{tab:ablation_arch}
\setlength{\tabcolsep}{2pt}
\renewcommand{\arraystretch}{1.3}
\begin{tabular}{l|cc|cc|cc|cc}
\multirow{2}{*}{Model} & \multicolumn{2}{c|}{Vision} & \multicolumn{2}{c|}{LLM} & \multicolumn{2}{c|}{CLS} & \multicolumn{2}{c}{MLLM} \\ \cline{2-9}
& \multicolumn{2}{c|}{CLIP / SigLIP} & \multicolumn{2}{c|}{Vicuna / Qwen2} & Gen. & Fgr. & Avg.$^*$ & Avg.$^\dag$ \\
\whline{1pt}
L-CV & ~~~~\cmark~~~ &  & ~~~~\cmark~~~~ &  & 75.3 & 47.1 & 54.9 & 42.8  \\ 
L-SV &  &\cmark & \cmark &   & 78.8 & 43.6 & 53.1 & 41.4 \\ 
L-CQ & \cmark &  &  & \cmark & 86.0 & 61.5 & 58.9 & 47.8 \\ 
L-SQ &  & \cmark &  & \cmark & 91.3 & 70.2 & 60.2 & 49.9 \\ 
\end{tabular}
\end{table}

\subsection{Network Architecture}
\label{sec:what makes good clser-net}
\noindent\textbf{Settings.}
In our investigation of MLLMs' network architecture, we explore various combinations of vision towers and LLMs, as illustrated in Table~\ref{tab:ablation_arch}. 
The training protocol is maintained as the standard two-stage LLaVA-1.5 \cite{llava}, utilizing the original LCS-558K data for Stage-1 pre-training and LLaVA-665K data for Stage-2 fine-tuning.
We assess the foundational classification performance across $10$ benchmarks, which are indicative of the model's ability to identify both general and fine-grained objects. 
Additionally, we also report performance metrics on $10$ widely recognized MLLM benchmarks to verify the efficacy of the modifications introduced.

\noindent\textbf{Analyses.}
As presented in Table~\ref{tab:ablation_arch}, we evaluate different combinations between vision towers ({\em e.g.}, OpenAI CLIP \cite{clip} and SigLIP \cite{siglip}) and LLMs ({\em e.g.}, Vicuna-1.5 \cite{zheng2023judging} and Qwen2 \cite{yang2024qwen2}).
We can observe that with Vicuna-1.5 as LLM, the modification by replacing CLIP with SigLIP (L-CV {\em v.s.} L-SV) results in a marginal decrease in MLLM performance.
Conversely, substantial improvements are observed in the L-CQ configuration, combining the CLIP vision tower with Qwen2.
This integration significantly enhances both classification and MLLM metrics (L-CV {\em v.s.} L-CQ), highlighting the influential role of Qwen2 in augmenting MLLM capabilities.
Besides, further progress (L-CQ {\em v.s.} L-SQ) can be noted when employing SigLIP in conjunction with Qwen2, particularly in fine-grained classification by $+8.7\%$, indicating the synergy between Qwen2 and specific vision towers can yield considerable advantages.

In conclusion, \textbf{the ablation on network architecture emphasizes the crucial role of the Qwen2 as LLM} in enhancing the capability across both image classification and more complex MLLM tasks.

\begin{table}[tp]
\centering
\small
\caption{\textbf{Ablation on the selection of training data.} 
In the `Model' column, `OV-' represents LLaVA-OV training protocol.
By default, LLaVA-665K (665K) is used during training, except for those models with `.' mark. `Mid', `SI' and `OV' are Mid-4M, SI-3.2M and OV-1.6M in LLaVA-OV.
IMN is the transformed ImageNet dataset used in \cite{VLMClassifier}. Other notations are same as Table~\ref{tab:ablation_arch}.}
\label{tab:ablation_data}
\setlength{\tabcolsep}{3pt}
\renewcommand{\arraystretch}{1.3}
\begin{tabular}{l|c|cc|c|cc|cc}
\multirow{2}{*}{Model} & \multicolumn{4}{c|}{Dataset} & \multicolumn{2}{c|}{CLS} & \multicolumn{2}{c}{MLLM} \\ \cline{2-9}
& Mid & \multicolumn{2}{c|}{~665K~~/~~SI} & OV & Gen. & Fgr. & Avg.$^*$ & Avg.$^\dag$ \\
\whline{1pt}
L-CV &  & \cmark &  &  & 75.3 & 47.1 & 54.9 & 42.8  \\ 

L-CV.IMN &  & \cmark+IMN &  &  & 68.6 & 36.4 & 54.0 & 41.5  \\ \hline
OV-SQ &  & \cmark &  &  & 91.7 & 73.2 & 61.6 & 54.2 \\ 
OV-SQ.Mid & \cmark & \cmark &  &  & 89.8 & 66.6 & 63.1 & 64.6 \\ 
OV-SQ.SI & \cmark &  & \cmark &  & 93.7 & 77.8 & 65.7 & 85.0  \\
OV-SQ.OV & \cmark &  & \cmark & \cmark & 94.1 & 78.6 & 65.5 & 86.8 \\  
\end{tabular}
\end{table}

\subsection{Training Data}
\label{sec:what makes good clser-data}

\noindent\textbf{Settings.}
In terms of training data, our primary focus is on the data selection
within Stage-1.5 for LLaVA-OV and Stage-2 for both LLaVA-1.5 and LLaVA-OV.
Specifically, for LLaVA-1.5, we utilize the standard LLaVA-1.5 \cite{llava} framework with the combination of CLIP \cite{clip} and Vicuna-1.5 \cite{zheng2023judging}, as well as the identical training protocol.
In this part, we examine whether directly incorporating domain-specific data (such as using ImageNet \cite{deng2009imagenet} in \cite{VLMClassifier}) enhances classification performance.
On the other hand, for LLaVA-OV, we adopt the official LLaVA-OV \cite{llava-ov} framework and the training protocol, with SigLIP \cite{siglip} and Qwen2 \cite{yang2024qwen2}, to investigate the efficacy of more diverse training data.

\noindent\textbf{Analyses.}
The detailed results are shown in Table~\ref{tab:ablation_data}.
Following the proposal in \cite{VLMClassifier}, we first evaluate directly incorporating ImageNet \cite{deng2009imagenet} data into the supervised fine-tuning in Stage-2 of LLaVA-1.5 \cite{llava}, resulting in poor classification and MLLM performances (such as $75.3\%$ of L-CV {\em v.s.} $68.6\%$ of L-CV.IMN in the general classification) due to overfitting.
\textbf{This demonstrates the potential drawbacks of relying on specialized data}.
Besides, when comparing OV-SQ and OV-SQ.Mid which additionally uses Mid-4M for the vision-language alignment \cite{llava-ov}, we find that despite the improvement on MLLM benchmark, it experiences degradation on both general and fine-grained classification. 
We conjecture this is because Mid-4M is primarily collected for injecting knowledge of general MLLM ability, such as document, OCR, Chinese and language understanding.
Moreover, building upon the Mid-4M alignment, one can see that OV-SQ.SI, which integrates SI-3.2M rather than LLaVA-665K in fine-tuning, achieves impressive results by improving $+3.9\%$ and $+11.2\%$ accuracy in general and fine-grained classification (OV-SQ.Mid {\em v.s.} OV-SQ.SI).
Similar trends by further including OV-1.6M within training can be observed.
Considering both SI-3.2M and OV-1.6M have much more balanced and diverse data sources than either ImageNet or LLaVA-665K, we \textbf{highlight the robustness and consistency of selecting diverse and extensive training data}.

\subsection{Training Recipe}
\label{sec:what makes good clser-recipe}

\begin{table}[tp]
\centering
\small
\caption{\textbf{Ablation on the strategy of training recipe.}
Here, `TV' and `AR' following the `.' mark denote tunable vision tower and image any resolution strategy \cite{llava-ov}, respectively.
Other notations are same as Table~\ref{tab:ablation_arch}~\&~\ref{tab:ablation_data}.
}
\label{tab:ablation_trainable}
\setlength{\tabcolsep}{2.6pt}
\renewcommand{\arraystretch}{1.3}
\begin{tabular}{l|c|c|cc|cc}
\multirow{2}{*}{Model} & \multicolumn{2}{c|}{Stage-2} & \multicolumn{2}{c|}{CLS} & \multicolumn{2}{c}{MLLM} \\ \cline{2-7}
& Tune Vision & AnyRes.\cite{llava-ov} & Gen. & Fgr. & Avg.$^*$ & Avg.$^\dag$ \\
\whline{1pt}

L-SV &  &  & 78.8 & 43.6 & 53.1 & 41.4 \\  % B
L-SV.TV & \cmark &  & 78.9 & 39.7 & 51.9 & 39.8 \\  
L-SV.AR &  & \cmark & 58.3 & 29.6 & 44.7 & 38.5 \\  
OV-SV & \cmark & \cmark & 70.3 & 32.6 & 45.0 & 38.5 \\ \hline

L-SQ &  &  & 91.3 & 70.2 & 60.2 & 49.9 \\ % D
L-SQ.TV & \cmark &  & 91.1 & 70.5 &61.0  & 49.7  \\ 
L-SQ.AR &  & \cmark & 91.6 & 73.4 &60.6  &53.7  \\ 
OV-SQ & \cmark & \cmark & 91.7 & 73.2 & 61.6 & 54.2 \\ 
\end{tabular}
\end{table}

\noindent\textbf{Settings.}
To explore the effect of the training recipe, we conduct ablations about the tunability of the vision tower and image high-resolution strategy, both adopted in LLaVA-OV \cite{llava-ov}.
Taking SigLIP \cite{siglip} as the vision tower, we consider two different LLM combinations, {\em i.e.}, Vicuna-1.5 \cite{zheng2023judging} and Qwen2 \cite{yang2024qwen2}.
Starting from the vanilla LLaVA-1.5 \cite{llava} training protocol, we gradually add the tunable vision tower and image any resolution strategy, ultimately aligning with the training protocol of LLaVA-OV.

\noindent\textbf{Analyses.}
Table~\ref{tab:ablation_trainable} gives the details of ablation.
One can see that when utilizing the image any resolution strategy (L-SV.AR), there is a precipitous decline in classification and MLLM benchmarks. 
This is due to the extra high-resolution image tokens with text tokens exceeding the maximum token length $4,096$ of Vicuna-1.5 \cite{zheng2023judging}, leading to the token truncation.
On the contrary, when transitioning from Vicuna-1.5 to Qwen2 \cite{yang2024qwen2} with a larger maximum token length of $32,768$, the incorporation of image any resolution has an immediate positive impact, bringing $+3.2\%$ accuracy gains for fine-grained classification.
Moreover, making the parameters in the vision tower tunable does not yield significant performance improvements (L-SQ {\em v.s.} L-SQ.TV).
To sum up, \textbf{using a tunable vision tower does not result in any significant performance gains, while the higher image high-resolution strategy requires a larger maximum token length}.

\section{Why Do the LLM and Data Matter?}
Based on the above analyses, the choice of LLM and diverse training data appear to be pivotal in improving the image classification capability of MLLMs.
To understand the rationale, we further explore the potential reasons from the aspects of LLM and training data.

\subsection{LLM - Better Conceptual Knowledge Transfer}

\begin{table}[tp]
\centering
\small
\caption{\textbf{Influence of adopting better LLM.} We report the accuracy on the entire ImageNet \cite{deng2009imagenet} and the worst three \texttt{classes} for Vicuna-1.5 \cite{zheng2023judging}.
$\bigtriangleup$ stands for the difference between LLM and MLLM models adopting Vicuna-1.5 and Qwen2 \cite{yang2024qwen2}.}
\label{tab:analysis_llm}
\setlength{\tabcolsep}{1.5pt}
\renewcommand{\arraystretch}{1.3}
\begin{tabular}{l|cc|cc|cc|cc}
\multirow{2}{*}{Model} & \multicolumn{2}{c|}{IMN} & \multicolumn{2}{c|}{\texttt{sorrel}} & \multicolumn{2}{c|}{\texttt{cairn}} & \multicolumn{2}{c}{\texttt{borzoi}} \\ \cline{2-9}
& LLM & MLLM & LLM & MLLM & LLM & MLLM & LLM & MLLM \\
\whline{1pt}
Vicuna & 70.7 & 76.3 & 16.0 & 20.0 &  6.0 &  8.0 & 14.0 & 20.0 \\ \hline
Qwen2  & 90.7 & 94.1 & 78.0 & 98.0 & 62.0 & 80.0 & 72.0 & 92.0 \\ 
\rowcolor{lightgray!20}
$\bigtriangleup$ & 20.0 & 17.8 & 62.0 & 78.0 & 56.0 & 72.0 & 58.0 & 72.0
\end{tabular}
\vspace{-5mm}
\end{table}

\noindent\textbf{Settings.}
As shown in Table~\ref{tab:ablation_arch}, replacing the LLM from Vicuna-1.5 \cite{zheng2023judging} to Qwen2 \cite{yang2024qwen2} (L-CQ \emph{vs.} L-CV) brings significant improvements in both classification and MLLM benchmarks.
We attribute the main cause to LLM's knowledge and capability obtained from its pre-training, to understand and distinguish different concepts.
To verify our assumption, we conduct experiments to compare the discriminative ability of vanilla Vicuna-1.5 and Qwen2
on category concepts within the ImageNet \cite{deng2009imagenet} dataset.

Specifically, for each image in the validation set of ImageNet, we utilize the powerful and public InternVL2-Llama3-76B \footnote{https://huggingface.co/OpenGVLab/InternVL2-Llama3-76B} as the privileged MLLM to identify the object class and then generate a justification accordingly.
We require the justification to cover as comprehensive aspects as possible, including target-object appearance, context or environment, function or behavior and other relevant information.
Besides, we also prohibit disclosing class names or related synonyms within the generation to prevent any leakage of the target class.
Detailed prompt is provided in the supplementary material.
Finally, these justifications are provided to vanilla Vicuna-1.5 and Qwen2 
, prompting them to choose a specific class from other confusing candidates, following the same protocol in Section~\ref{sec:cls_eval_prot}.
For instance, a sample question with candidate options can be as follows: 
\begin{quotation}
\noindent Here is a description of someone object.
\textit{\textbf{\textcolor{red}{\textless description \textgreater}}
Can you tell me what object it describes? 
Answer with the option’s letter from the given choices directly.\textbackslash n
A. \textbf{\textcolor{blue}{\textless class name A\textgreater}}\textbackslash n
B. \textbf{\textcolor{blue}{\textless class name B\textgreater}}\textbackslash n
$\cdots$
Z. \textbf{\textcolor{blue}{\textless class name Z\textgreater}} 
}
\end{quotation}
where \textit{\textcolor{red}{description}} is the generated justifications by the privileged MLLM for the correct class and \textit{\textcolor{blue}{class name A ... Z}} are the candidate class names.

\noindent\textbf{Analyses.}
As listed in Table~\ref{tab:analysis_llm}, we compare vanilla LLMs adopting Vicuna-1.5 \cite{zheng2023judging} and Qwen2 \cite{yang2024qwen2} on the entire ImageNet \cite{deng2009imagenet} dataset, as well as the worst specific classes within the Vicuna-1.5's evaluation.
One can see that Qwen2 achieves better accuracy ($90.7\%$ {\em v.s.} $70.7\%$) than Vicuna-1.5 on ImageNet.
Besides, in the worst specific classes ({\em e.g.}, sorrel, cairn and borzoi) of Vicuna-1.5, Qwen2 obtains much better performance with over $+56.0\%$ gains, indicating that Qwen2 has richer conceptual knowledge than Vicuna-1.5.
Moreover, we also review the classification performance of MLLMs that adopt Vicuna-1.5 and Qwen2 in Table~\ref{tab:analysis_llm}.
Similar trends can be observed, which demonstrates that \textbf{Qwen2 enables a better conceptual knowledge transfer into the MLLM framework}, thereby resulting in better classification performance.

\subsection{Data - Enhanced Exposure to Target Concepts}

\begin{table}[tp]
\centering
\small
\caption{\textbf{Influence of including more target-concept-related data.}
`F' and `NF' represent food-related and non-related subsets.
$\bigtriangleup$ stands for the difference with `L-CV'.
`SI.Food' is the food-related data gathered from SI-3.2M.
Food-related subsets and $\bigtriangleup$ values are marked in \textbf{bold}.}
\label{tab:analysis_data}
\setlength{\tabcolsep}{2pt}
\renewcommand{\arraystretch}{1.3}
\begin{tabular}{l|r|rr|rr|rr|rr}
\multirow{2}{*}{Model} & Food & \multicolumn{2}{c|}{IMN} & \multicolumn{2}{c|}{IMNv2} & \multicolumn{2}{c|}{CLS} & \multicolumn{2}{c}{MLLM} \\ \cline{2-10}
& \textbf{All} & \textbf{F} & NF & \textbf{F} & NF & Gen. & Fgr. & Avg.$^*$ & Avg.$^\dag$ \\
\whline{1pt}
L-CV   & 71.0 & 76.6 & 76.3 & 79.8 & 78.3 & 75.3 & 47.1 & 54.9 & 42.8 \\ \hline 
+ SI.Food & 77.2 & 82.7 & 79.6 & 85.5 & 81.6 & 77.8 & 47.8 & 55.4 & 45.7 \\
\rowcolor{lightgray!20}
$\bigtriangleup$ & \textbf{6.2} & \textbf{6.1} & 3.3 & \textbf{5.7} & 3.3 & 2.5 & 0.7 & 0.5 & 2.9 \\ 
\end{tabular}
\end{table}

\noindent\textbf{Settings.}
Table~\ref{tab:ablation_data} shows replacing LLaVA-665K \cite{llava} with SI-3.2M \cite{llava-ov} as the training data improves the classification performance by a large margin.
We hypothesize this improvement is primarily due to the introduction of more domain-specific knowledge during the supervised fine-tuning in Stage-2.
Thus, we propose augmenting the Stage-2 with additional domain-specific training data to assess whether the benchmark of the target domain improves.

Specifically, we select LLaVA-1.5 \cite{llava} as our baseline model and take `food' as the specific domain to verify our hypothesis.
This is mainly considering that the corresponding Food101 \cite{bossard2014food} benchmark contains i) the largest amount of samples for a more robust evaluation and ii) less conceptual ambiguity between different classes ({\em e.g.}, the term `rose' in Flower102 \cite{flower102} referring to both color and flower types).
Based on SI-3.2M, we gather food-related data by tokenizing all the samples and then selecting those whose conversations contain any classes in Food101.
By filtering out the overlapping data within LLaVA-665K, we append the remaining ones to create an enhanced domain-specific dataset, ultimately resulting in $(665+49)$K samples.
Following the training recipe of LLaVA-1.5, we re-perform the supervised fine-tuning in Stage-2 with the new dataset.

\noindent\textbf{Analyses.}
As presented in Table~\ref{tab:analysis_data}, we report the accuracies on both food-related and non-related subsets, respectively.
More details about the food-related classes can be found in the supplementary material.
One can see that incorporating our proposed enhanced domain-specific dataset (denoted as `+SI.Food') leads to a notable performance improvement, particularly in food-related classes under ImageNet \cite{deng2009imagenet}, ImageNetv2 \cite{imagenetv2} and Food101 \cite{bossard2014food}.
For example, on the Food101 dataset, we observe a significant increase (L-CV {\em v.s.} +SI.Food) by $+6.2\%$ than the LLaVA-1.5 baseline \cite{llava}.
Similarly, for ImageNet and ImageNetv2, the improvements also exceed $+6.1\%$ and $+5.7\%$, respectively.
Moreover, due to the diversity and balance of SI-3.2M data, the newly-combined enhanced domain-specific dataset can also slightly improve the performance of food non-related subsets, as well as the MLLM benchmarks.
Building upon the above analyses, we conclude that \textbf{the enhanced exposure of target concepts in diverse training data benefits the classification capability of MLLM}.

\section{Conclusion}
In this paper, we revisit MLLMs with an in-depth analysis of image classification.
We build a comprehensive image classification benchmark and find that the most recent MLLMs have caught up with CLIP-style vision-language models.
Through the analysis of network architecture, data selection and training recipe, we attribute the potential reasons to conceptual knowledge transfer in LLMs and enhanced exposure of target concepts in training data.

\noindent\textbf{Limitations and future works.}
Due to the limited computing resources, this paper typically focuses on MLLMs with sizes less than $10$B. For the image classification evaluation of larger models, we leave it to future work. Besides, the MLLM evaluation of other conventional tasks, such as object detection, is also worth a deeper study.

{
    \small
    \bibliographystyle{ieeenat_fullname}
    \bibliography{main}
}

\clearpage

\begin{appendices}
\section*{Appendix}
In this appendix, we first discuss the \textit{potential negative societal impacts} (refer to Section~\ref{impacts}) that may arise in practical scenarios.
Then, we perform an in-depth exploration of \textit{MLLM classification evaluation} (delineated in Section~\ref{MLLM_cls_eval}), including the formulation, influence of option numbers, etc.
Next, we detail the prompt and ablation of \textit{privileged MLLM} (outlined in Section~\ref{priv_MLLM}) used for the evaluation of LLMs.
Lastly, we give the \textit{food-related subset selection} (explicated in Section~\ref{food_select}) in our evaluation of Table~9.

\section{Broader impacts}
\label{impacts}
The evaluation of MLLMs, while aimed at establishing reliable performance benchmarks, may inadvertently lead to incomplete or oversimplified conclusions. For instance, although MLLMs demonstrate strong classification performance, numerous critical scenarios, such as healthcare or autonomous driving, remain unexplored in this paper. Deploying MLLMs in such domains without additional evaluation could lead to inappropriate applications, resulting in suboptimal or even adverse outcomes. These risks need careful consideration and extensive validation when applying MLLMs to practical classification tasks.

\section{MLLM classification evaluation}
\label{MLLM_cls_eval}
In this section, we conduct additional explorations of the MLLM classification evaluation.
First, we investigate different formulations of MLLM evaluation, such as free-form answering.
Then, an ablation is provided to assess the impact of different numbers of options. 
Finally, we evaluate different strategies for selecting question options.

\subsection{Formulations of evaluation}
Early MLLM benchmarks, such as LAMM \cite{yin2024lamm} and LVLM-eHub \cite{xu2023lvlm}, adopt evaluation samples with free-form answering.
Following this formulation, we directly ask MLLMs to give the class names from the input \textit{\textcolor{red}{image}}:
\begin{tcolorbox}[title=Free-form Answering Prompt]
\textcolor{red}{$<$image$>$}
Please identify the main object in the given image and output the category name.
\end{tcolorbox}

\noindent However, due to the preference within training data, MLLMs such as LLaVA-OV \cite{llava-ov} tend to answer common categories while missing the target rare class, such as identifying a `person' rather than the clothing `bikini' in ImageNet \cite{deng2009imagenet}.
To address this issue, we further attempt to provide a candidate list of all \textcolor{blue}{class names} within the prompt, as follows:
\begin{tcolorbox}[title=Free-form Answering Prompt with Candidates]
\textcolor{red}{$<$image$>$}
Please identify the main object in the given image and select the category name from the following list.\\
$[$ \textcolor{blue}{$<$class name 1$>$}, \textcolor{blue}{$<$class name 2$>$} , ...$]$
\end{tcolorbox}

\noindent However, including all class names within the prompt significantly increases the token length (exceeding 10,000), often surpassing the maximum token limit that MLLMs can handle. For instance, MLLMs utilizing Vicuna-1.5 \cite{zheng2023judging} as the LLM, which has a limit of less than 4,096 tokens, are unable to process such lengthy inputs.
Besides, for MLLMs with extensive context processing capabilities, such as LLaVA-OV \cite{llava-ov} and Qwen2-VL \cite{qwen2vl}, which adopt Qwen2 \cite{yang2024qwen2} as the LLM, we observe that these models often generate lengthy responses, with occasional instances of hallucination.
In this context, evaluating the quality of such free-form outputs typically requires either human annotators or GPT-based assistants, as similarly noted in \cite{li2023seed}.

Therefore, considering the limitations of free-form answering and following \cite{li2023seed}, we adopt multiple-choice questions (MCQ) for classification evaluation with a clear option output format, providing an objective and efficient evaluation formulation.

\subsection{Number of options}
Different from the commonly used $4$-choice questions in \cite{li2023seed,liu2024mmbench}, in our classification benchmark, each question owns $26$ candidate options by default.
Based on the ImageNet \cite{deng2009imagenet} and Food101 \cite{bossard2014food} datasets, this subsection further ablates the influence of different numbers of options for Qwen2-VL \cite{qwen2vl}.
The results are shown in the following table.
\begin{table}[h]
\vspace{-3mm}
\centering
\small
\setlength{\tabcolsep}{4pt}
\renewcommand{\arraystretch}{1.1}
\begin{tabular}{c|cc}
\#option & ImageNet \cite{deng2009imagenet} & Food101 \cite{bossard2014food} \\
\whline{1pt}
4  & 99.5 & 99.3 \\
8  & 99.1 & 98.6 \\
16 & 98.3 & 97.5 \\ \hline
26 & 97.4 & 96.2
\vspace{-3mm}
\end{tabular}
\end{table}

\noindent We can see that increasing the number of options raises the difficulty of the evaluation questions, resulting in slight performance decreases. Therefore, we use $26$ options, labeled from `A' to `Z', to enhance the challenge.

\subsection{Strategy of negative option selection}
In Section~\textcolor{red}{3.2} of our paper, we generate incorrect options for each multiple-choice question through random selection. In this subsection, we explore a different strategy, {\em i.e.}, utilizing the top-25 most similar class names identified by BERT \cite{kenton2019bert} as the candidate options.
Concretely, for each class within the dataset, we first calculate its vector representation using a pre-trained BERT and then select the top-25 most similar ones using the cosine similarity metric.
The following table reports the corresponding influence for OpenAI CLIP \cite{clip}, SigLIP \cite{siglip} and Qwen2-VL \cite{qwen2vl}.
\begin{table}[h]
\vspace{-3mm}
\centering
\small
\setlength{\tabcolsep}{6pt}
\renewcommand{\arraystretch}{1.3}
\begin{tabular}{l|cc}
Model & ImageNet \cite{deng2009imagenet} & Food101 \cite{bossard2014food} \\
\whline{1pt}
\textcolor{gray}{OpenAI CLIP} & \textcolor{gray}{94.1} & \textcolor{gray}{97.1}\\
$\boldsymbol{\bigtriangleup}$ & \textbf{-2.2} & \textbf{-0.8}\\
\textcolor{gray}{SigLIP}      & \textcolor{gray}{96.2} & \textcolor{gray}{98.5} \\ 
$\boldsymbol{\bigtriangleup}$ & \textbf{-1.5} & \textbf{-0.2}\\ \hline 
\textcolor{gray}{Qwen2-VL}    & \textcolor{gray}{95.1} & \textcolor{gray}{95.7}\\
$\boldsymbol{\bigtriangleup}$ & \textbf{-2.3} & \textbf{-0.5} 
\vspace{-3mm}
\end{tabular}
\end{table}

\noindent It is evident that when more confusing classes are used as incorrect options instead of the randomly selected ones, all models show a similar and slight decrease in performance.
However, recently proposed MLLMs, such as Qwen2-VL, still achieve classification results comparable to those of CLIP-style models, {\em e.g.}, CLIP and SigLIP.

\section{Privileged MLLM}
\label{priv_MLLM}
In Section~\textcolor{red}{5.1} of our paper, we leverage a privileged MLLM ({\em i.e.}, InternVL2-Llama3-76B \cite{internvl,internvl1.5}) to help the evaluation of LLM’s knowledge and capability to understand and distinguish different concepts, including Vicuna-1.5 \cite{zheng2023judging} and Qwen2 \cite{yang2024qwen2}.
This section gives the details about the ablation of using different MLLMs as the privileged model, as well as our adopted prompt in generating the justifications from the privileged MLLM.

\subsection{Different MLLMs as the privileged model}
In our paper, we adopt InternVL2-Llama3-76B\footnote{https://huggingface.co/OpenGVLab/InternVL2-Llama3-76B} as the privileged model.
Here, we ablate the effect of adopting different MLLMs, {\em e.g.}, changing to the powerful and public Qwen2-VL-72B\footnote{https://huggingface.co/Qwen/Qwen2-VL-72B-Instruct}.
The following table gives the results:
\begin{table}[h]
\vspace{-3mm}
\centering
\small
\setlength{\tabcolsep}{3pt}
\renewcommand{\arraystretch}{1.3}
\begin{tabular}{l|cc}
MLLM & Vicuna-1.5 \cite{zheng2023judging} & Qwen2 \cite{yang2024qwen2} \\
\whline{1pt}
InternVL2-Llama3-76B  & 70.7 & 90.7\\ \hline
Qwen2-VL-72B          & 75.4 & 91.5 
\end{tabular}
\vspace{-3mm}
\end{table}

\noindent We observe that when switching the privileged model to Qwen2-VL-72B, Qwen2 still achieves a much better performance than Vicuna-1.5 ($91.5\%$ {\em v.s.} $75.4\%$), demonstrating Qwen2 has better conceptual knowledge than Vicuna-1.5, consist with the conclusion in our main paper.

\subsection{Prompt for privileged MLLM justifications}
To cover as comprehensive aspects as possible, we utilize the following prompt for the justification generation, where \textit{\textcolor{red}{image}} represents the currently handled image and \textit{\textcolor{blue}{class name}} is the ground truth class label of the current image:
\begin{tcolorbox}[title=Justification Generation Prompt]
\textcolor{red}{$<$image$>$}
You will be given a \textcolor{blue}{$<$class name$>$} image. Your task is to identify the \textcolor{blue}{$<$class name$>$} in it and summarize the reason for this judgment. You should focus on several aspects, including: \\
1. \textbf{Appearance}: Describe the object's shape, color, texture, and any distinct features. \\
2. \textbf{Context or environment}: Where is this object commonly found or used? Describe its typical environment or surroundings. \\
3. \textbf{Function or behavior}: Explain how this object is typically used or behaves (if it's an animal or device, for example). \\
4. \textbf{Any other relevant information}: Mention any notable facts or characteristics that help define this object class. \\
\textbf{NOTE}: \\
1. Keep the reason concise, with one or two sentences for each aspect. \\
2. Do not use the category or synonym names of \textcolor{blue}{$<$class name$>$} in your answer, instead using `it'. \\
3. Just output the reason.
\end{tcolorbox}

\section{Food-related subset selection}
\label{food_select}
In Section~\textcolor{red}{5.2}, for a more comprehensive analysis, we manually categorize the classes in ImageNet \cite{deng2009imagenet} and ImageNetv2 \cite{imagenetv2} into food-related and non-related subsets. 
Specifically, to determine the class attribution, we enumerate the classes and employ GPT-4 \cite{achiam2023gpt} to classify each of them. 
The corresponding prompt is as follows, where \textcolor{blue}{\textit{class name}} is one of the candidate class names:
\begin{tcolorbox}[title=Food-related Categorizing Prompt]
Answer the yes or no question below.\\
Is \textcolor{blue}{$<$class name$>$} commonly considered a type of food or dish that will be eaten by people?\\
Output only one choice [ YES / NO ].
\end{tcolorbox}

\noindent Finally, there are $83$ classes belonging to the food-related subset in the ImageNet dataset, while $86$ for the ImageNetv2 dataset, as listed in the following table.
\begin{table*}[t]
\vspace{-15cm}
\centering
\small
\setlength{\tabcolsep}{2pt}
\begin{tabular}{l|p{0.795\linewidth}}
Dataset & Classes in food-related subset\\
\whline{1pt}
\hline
ImageNet & 
acorn squash, american lobster, artichoke, bagel, bakery, banana, barracouta, bell pepper, bison, black grouse, bolete, broccoli, burrito, butternut squash, carbonara, cauliflower, cheeseburger, chocolate sauce, cock, coho, conch, confectionery, consomme, corn, crayfish, cucumber, custard apple, dough, drumstick, dungeness crab, eel, fig, french loaf, gar, goose, granny smith, guacamole, hare, head cabbage, hen, hog, honeycomb, hotdog, ice cream, ice lolly, jackfruit, jellyfish, king crab, lemon, lionfish, mashed potato, meat loaf, mushroom, orange, partridge, pineapple, pizza, pomegranate, potpie, prairie chicken, pretzel, ptarmigan, puffer, quail, rock crab, ruffed grouse, scorpion, sea anemone, sea cucumber, sea slug, sea urchin, snail, sorrel, spaghetti squash, spiny lobster, strawberry, sturgeon, tench, terrapin, trifle, wild boar, wing, zucchini\\ \hline
ImageNetv2 & 
acorn squash, american bullfrog, american lobster, artichoke, bagel, baguette, bakery, banana, bell pepper, bison, black grouse, bolete, broccoli, burrito, butternut squash, cabbage, carbonara, cauliflower, cheeseburger, cherimoya (custard apple), chocolate syrup, chow chow, conch, consomme, corn, corn cob, cottontail rabbit, crayfish, cricket insect, cucumber, dough, drumstick, duck, dungeness crab, eel, fig, goose, granny smith apple, guacamole, hare, hen, hen of the woods mushroom, honeycomb, hot dog, ice cream, jackfruit, jellyfish, lemon, lionfish, mashed potatoes, meatloaf, mushroom, orange, partridge, pig, pineapple, pizza, pomegranate, popsicle, pot pie, pretzel, ptarmigan, pufferfish, quail, ram (adult male sheep), red king crab, rock crab, rose hip, ruffed grouse, scorpion, sea anemone, sea cucumber, sea slug, sea urchin, silver salmon, snail, snoek fish, spaghetti squash, spiny lobster, strawberry, sturgeon, tench, terrapin, trifle, wild boar, zucchini \\ 
\end{tabular}
\vspace{-3mm}
\end{table*}

\end{appendices}

\end{document}